\documentclass{article}

     \PassOptionsToPackage{numbers, compress}{natbib}

\usepackage[final]{lmca2020}

\usepackage[utf8]{inputenc} 
\usepackage[T1]{fontenc}
\usepackage{hyperref}       
\usepackage{url}            
\usepackage{booktabs}       
\usepackage{amsfonts}       
\usepackage{nicefrac}       
\usepackage{microtype}      
\usepackage{graphicx}
\usepackage{amsmath}

\title{Evaluating Curriculum Learning Strategies\\in Neural Combinatorial Optimization}

\author{%
Michal Lisicki\\
University of Guelph\\
Vector Institute for AI\\
\texttt{mlisicki@uoguelph.ca}
\And~Arash Afkanpour\\
Google \\
\texttt{arashaf@google.com}
\And~Graham W.~Taylor\\
University of Guelph\\
Vector Institute for AI\\
\texttt{gwtaylor@uoguelph.ca} }

\begin{document}

\maketitle

\begin{abstract}

Neural combinatorial optimization (NCO) aims at designing problem-independent
and efficient neural network-based strategies for solving combinatorial problems.
The field recently experienced growth by successfully adapting architectures
originally designed for
machine translation. Even though the results are promising, a large
gap still exists between NCO models and classic deterministic solvers, both in
terms of accuracy and efficiency. One of the drawbacks of current approaches
is the inefficiency of training on multiple problem sizes. Curriculum learning
strategies have been shown helpful in increasing performance in the multi-task
setting. In this work, we focus on designing a curriculum learning-based training
procedure that can help existing architectures achieve competitive performance
on a large range of problem sizes simultaneously. We provide a systematic
investigation of several training procedures and use the insights gained to
motivate application of a psychologically-inspired approach to improve upon
the classic curriculum method.

\end{abstract}

\section{Introduction}

Attention and sequence-to-sequence models have recently been shown
capable of approximating solutions to
combinatorial problems \citep{bello2016neural,kool2019attention,vinyals2015pointer}.
However, a large performance gap still exists between those models and the exact \citep{applegate2006concorde}
or heuristic \citep{christofides1976worst,helsgaun2017extension,karlin2020slightly,lin1973effective,serdjukov_extremal_1978} solvers.
So far, researchers have mostly focused on training separate models for
each problem size. The ability of a model to perform well across
a larger range of sizes simultaneously has been mostly overlooked.

Training and testing on individual problem sizes misleadingly diminishes the
performance gap between neural combinatorial
optimization (NCO) solutions and solvers. While
algorithms are often designed to provide bounded approximation
\cite{christofides1976worst,karlin2020slightly,serdjukov_extremal_1978}, NCO requires an ensemble of models to achieve
similar performance, which can be prohibitive in industrial applications~\cite{cook2006tsp}.

Curriculum learning (CL) is based on the premise that neural networks learn
faster from tasks gradually increasing in complexity
\citep{bengio2009curriculum,elman1993learning}. CL algorithms used
for scheduling tasks are called \textit{sampling strategies}. The na\"ive
strategy often leads to catastrophic forgetting and decline of efficiency.
Adaptive staircase \citep{leibo2018psychlab}, a strategy developed in psychophysics, incorporates rehearsal and
adjustment of difficulty to the current capacity, which were shown to significantly
improve performance in multi-task learning \citep{lange2019continual}.
In this work we assess CL as an effective training procedure to help a recent state-of-the-art
NCO model \citep{kool2019attention} achieve competitive performance on a range
of problem sizes simultaneously.\looseness=-1

We provide a novel application of the CL framework in NCO. Our assessment of the model's performance
on problem sizes other than it was trained on shows that the extent of knowledge that can be
transferred between tasks changes smoothly with problem size.
Investigation of several baseline strategies shows a clear benefit from rehearsal and
training on problems of increasing difficulty. These insights serve to motivate
a psychologically-inspired approach that improves upon classic CL.\@

\section{Background}

\textbf{Travelling Salesperson Problem.} The travelling salesperson problem (TSP) is
a combinatorial problem in which we look for the shortest path through $n$
cities, visiting each city exactly once, and return to the origin. The NCO
model we utilize was designed to operate on the Euclidean TSP,
which requires the distances between the cities to be defined using the Euclidean metric.
Imposing such a restriction allows for clear visualizations and direct comparison of the results 
with previous research.
The problem was shown to be NP-complete \citep{applegate2006concorde}, but in practice
we can find solutions within a reasonable time
frame, using highly optimized solvers, like Concorde \citep{applegate2006concorde}.

TSP can be formulated as a search for an optimal permutation of
input cities. Following the notation from \citep{kool2019attention},
let $\boldsymbol\pi = \{\pi_1,\ldots,\pi_n\}$, where each element
$\pi_i\in\{1,\ldots,n\}$ is a node index, such that $\pi_i\neq \pi_{j}\
\forall i\neq j$. Given a distance matrix $D$, where input elements $D_{ij}$
correspond to a cost of travel between a city $i$ and a city $j$, our objective is
to find a permutation resulting in a tour of minimal length
$L(\boldsymbol\pi) = \sum_{i=1}^n D_{\pi_i \pi_{i+1}}$.\footnote{The additional
element $\pi_{n+1}=\pi_1$ in the equation ensures the closure of the path.}

\textbf{Evaluation Metric.} Comparing the performance of combinatorial problems that
differ in complexity is challenging and often requires domain-specific
knowledge. A commonly used metric is the approximation ratio $R(x,y) = \frac{f(y)}{\text{OPT}(x)}$
between the score $f(y)$ of an approximate feasible solution $y$ to a problem instance $x$,
and the score of an optimal solution $\text{OPT}(x)$~\citep{kann1992approximability, khalil2017learning}.
Constant factor approximation algorithms are expected
to be no worse than a given ratio, independent of the problem size.
For TSP the objective $f(y) = L(\boldsymbol\pi)$.
Most of the previous work in NCO \citep{bello2016neural,kool2019attention,vinyals2015pointer}
uses a variant of the approximation ratio, called the \textit{optimality gap}
$\big(R(x,y)-1\big)$, and propose using a Concorde solution as a reference point in the ratio
\citep{bello2016neural,kool2019attention,vinyals2015pointer}.
In this work we calculate the optimality gap w.r.t.\ the Held-Karp (HK) lower bound~\citep{held1970traveling},
as it is independent of any software, well understood in the combinatorial optimization community \citep{christofides1979combinatorial}, and easy to implement.

\textbf{Neural Combinatorial Optimization.} Most of the recent advances in NCO are based on
sequence-to-sequence models \citep{bello2016neural,kaempfer2019learning,vinyals2015pointer},
attention models \citep{kool2019attention}, or graph neural networks\footnote{The authors of \citep{kool2019attention}
point out that their model is equivalent to the graph attention network proposed in
\cite{velivckovic2017graph}.} (GNN)
\citep{dwivedi2020benchmarking,joshi2020learning}.
Solutions inspired by successful applications in machine translation \citep{sutskever2014sequence}
do not carry forward directly to TSP, as the
the models depend on a fixed dictionary. Pointer Networks
\citep{vinyals2015pointer} train an attention mechanism to
point back to the input, which eschews the need for a separate
dictionary and allows for training on graphs of variable size.
Reinforcement learning was later shown to provide a more suitable specification
of the objective function for combinatorial problems than supervised learning
\citep{bello2016neural,kool2019attention}. The attention model (AM),
proposed by \citep{kool2019attention}, builds upon the empirical evidence that pure
attention-based models tend to outperform their recurrent neural network counterparts
on sequence prediction tasks \citep{devlin2018bert,vaswani2017attention},
mainly due to improvements in algorithmic efficiency and regularization effects.
While graph convolutional networks show better performance on individual problem sizes,
the attention-based methods provide comparable results with better generalization
\cite{joshi2019efficient,joshi2020learning}.

\textbf{Curriculum Learning.} Curriculum learning (CL) is a training procedure
that utilizes knowledge transfer between tasks of increasing difficulty and
devises an efficient task sampling strategy. While the difficulty of a problem
can be defined with respect to the capability of a model to solve it,
we impose a further restriction and assume a global ordering of tasks, with a direct
correspondence between the problem's size and its difficulty level.
CL attemps to exploit a neural network's
ability to facilitate learning solutions to difficult tasks with the knowledge of simpler ones.
Recently, there has been a lot of focus on algorithms that can automate the
process of designing curricula for specific applications, particularly those
stemming from reinforcement learning \citep{graves2017automated,portelas2020automatic}.

\textbf{Adaptive staircase.} Adaptive staircase is an improved CL strategy adapted
from psychophysics to deep reinforcement learning \citep{leibo2018psychlab}. At
each step, the strategy determines if it is either: ready to advance to a more difficult
task, backtrack, or train on a task at the current difficulty level. In the base
case, the agent is always trained on a task of the current difficulty level $t$.
The alternative is a probe case, where the task is sampled uniformly from a range
$[1,t]$. The base and probe cases are chosen equiprobably at each trial. The
level is incremented to $t+1$ when the agent achieves a certain performance
threshold $\alpha$ after observing $N_\text{base}$ base trials. Probe trials
contribute to training, but the agent is not evaluated on them. The number of
base trials per task can either be constant or can be proportional to the
current difficulty level: $N_\text{base}\propto t$.

\section{Experiments and discussion}

To test the effectiveness of CL in improving NCO performance on a range
of problem sizes, we performed two experiments.
We started by analyzing the ability of the AM model \citep{kool2019attention} to solve tasks on which it
was not trained. The results showed the limitations of knowledge transfer
between the models, which prompted our investigation of several different sampling
strategies: (i) single task, (ii) stochastically sampled range of tasks,
and (iii) two curriculum learning approaches. In all the experiments we
used REINFORCE with a greedy rollout baseline with a learning
rate $\eta=10^{-3}$. Training each of the methods for 100 epochs took $\approx 50$ hours
on 4 P100 GPUs, depending of the dominant size of the problem. Following
\citep{kool2019attention}, we initialized parameters with $\text{Uniform}(-1/\sqrt{d},1/\sqrt{d})$
for the input dimension $d$. Every epoch the network used 2,500
batches of 512 samples.\footnote{In case of adaptive staircase the number varied
per epoch. This is explained later in the section.} As a metric we chose the optimality gap,
reported as percentage, between
the average cost returned by the model and the HK lower bound estimate computed over
$10^4$ TSPs per problem size. For the smallest problem sizes (4--9) we
ran exhaustive search over $10^5$ TSPs to reduce the variability.\looseness=-1

\begin{figure}[ht]
\begin{minipage}[t]{0.48\textwidth}
\centering
\includegraphics[width=1.0\textwidth]{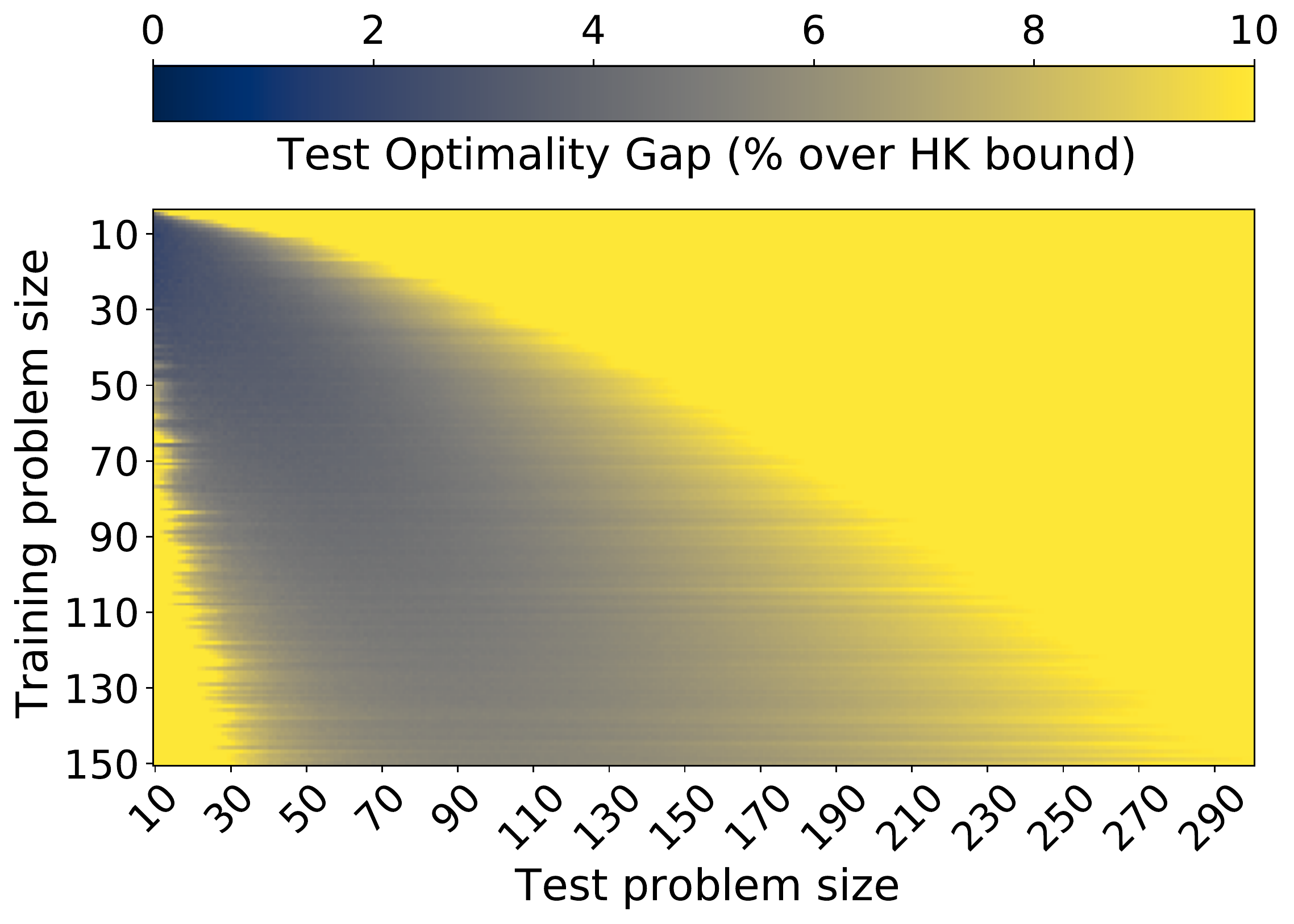}
\caption{Heat map of validation scores of models trained separately on problem sizes 4 to
150. The optimality gap is measured with respect to the HK lower bound and the threshold
caps it from the top at 10\%.}
\label{fig:performance_matrix}
\end{minipage}%
\hfill
\begin{minipage}[t]{0.48\textwidth}
\centering
\includegraphics[width=1.0\textwidth]{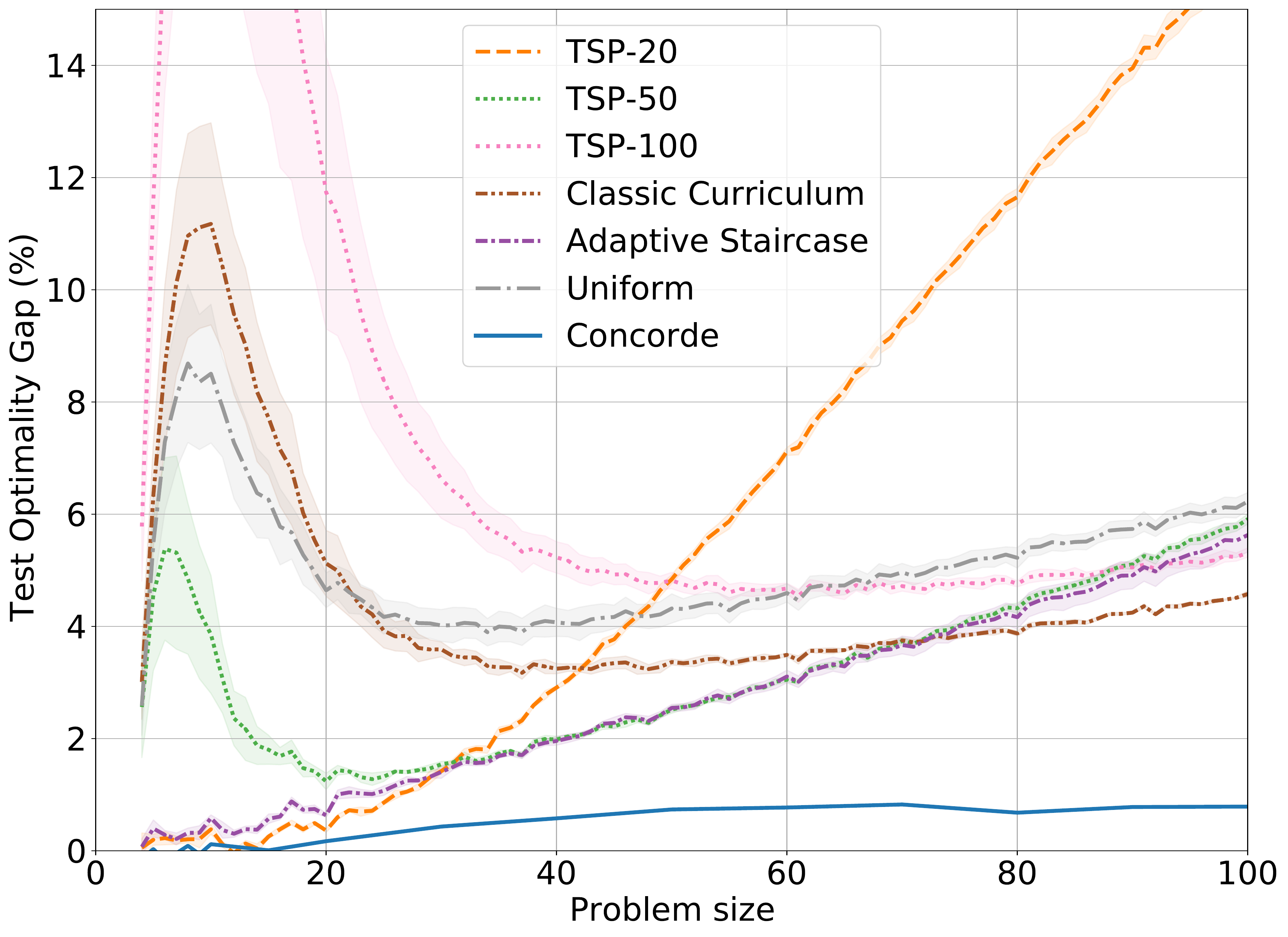}
\caption{The optimality gap of the final epoch performance of each of the
  models w.r.t.\ exhaustive search (sizes 4--9) and HK bound (sizes 10--150).
  The confidence intervals represent 2 standard errors over 5 runs.}
\label{fig:all_methods_opt_gap}
\end{minipage}
\vspace{-12px}
\end{figure}

To quantify the capability of the model to generalize to other tasks
we trained 147 models from TSP 4 to 150,
and tested them on sizes 10 to 300. HK lower bounds for this experiment
were extrapolated from the available representative set.
Each of the models was initialized with the parameters of a model pretrained in
a classic curriculum fashion, and subsequently trained for 20 epochs.
Fig.~\ref{fig:performance_matrix} shows the performance matrix, in a form of a heat map, where
each row corresponds to the performance of a model trained on a single task when tested on a range
of tasks from 10 to 300.

\begin{figure}[ht]
\centering
\includegraphics[width=1.0\textwidth]{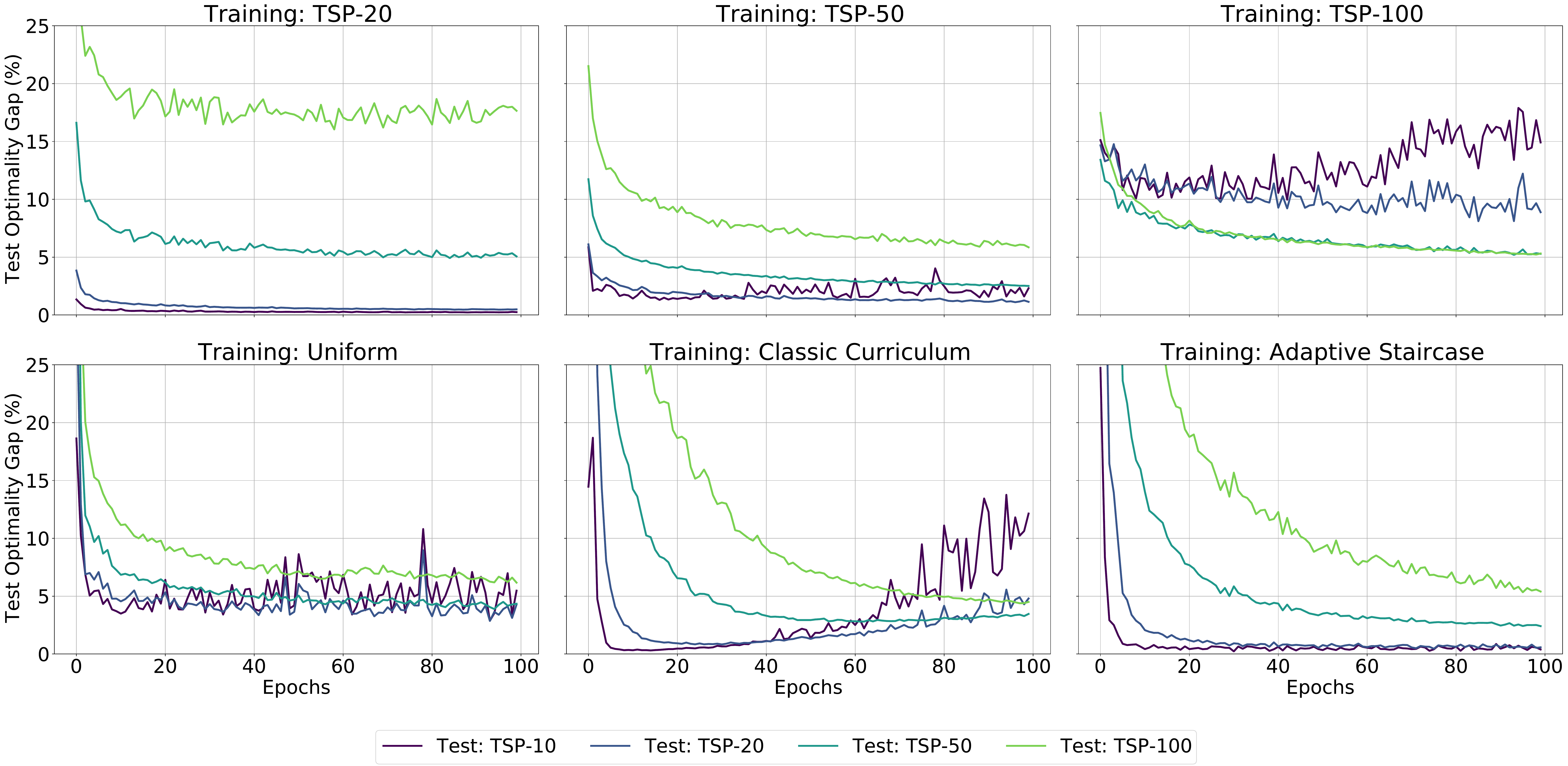}
\caption{For each sampling strategy, we follow the training procedure of the Attention Model
(AM) proposed in~\citep{kool2019attention}. The curves are color coded from dark
to bright in increasing level of task difficulty. Each subplot title denotes the
task schedule on which the model was trained. The ``Test: TSP-$N$'' curve represents
validation of the model on a task $N$ at every epoch.}
\label{fig:training_curves}
\end{figure}

In the second experiment, we compared six sampling strategies (Fig.~\ref{fig:training_curves}).
Following previous work \citep{bello2016neural,kool2019attention,vinyals2015pointer},
we first trained three models on fixed problem sizes 20, 50 and 100.
For the rest of the strategies, we trained the model on a set of TSPs from
4 (the minimal meaningful problem size) to 100.
We started with a baseline (uniform) stochastic model that selected
a task randomly at each epoch. Then we tested the two CL approaches ---
the classic approach of monotonically increasing problem sizes, and the adaptive staircase.

\addtocounter{footnote}{-1}

We start the discussion by observing that the performance of models trained on neighboring
tasks is correlated (Fig.~\ref{fig:all_methods_opt_gap}, Fig.~\ref{fig:performance_matrix}).
The figures show that the performance of different models can be
delineated by a skewed Gaussian centered at the task on which the model was trained.
The ability of the model to solve problems close in size suggests a large knowledge transfer,
which occurs with widening spread but smaller magnitude as the
problem size increases (Fig.~\ref{fig:performance_matrix}). Moreover,
a drift from smaller problem sizes implies that a successful sampling strategy
must be able to balance small and large problem sizes.\looseness=-1

Throughout the course of training (Fig.~\ref{fig:training_curves}), the models
that do not visit the previous problems along with the new ones show initial
improvement in performance, but as soon as the network reaches a certain epoch
the performance decays for the tasks that are further away. This shows the
limitation of knowledge transfer and exposes catastrophic forgetting.

The model trained with uniform sampling (Fig.~\ref{fig:all_methods_opt_gap})
is the closest in performance to a constant factor approximator, but its
optimality gap across all the problem sizes is larger than most of the other methods.
We conclude that balancing the problem sizes na\"ively does not
lead to an overall improvement in performance. Knowledge transfer discrepancies,
catastrophic forgetting and lack of efficiency prompted our
investigation of curriculum-based methods.

Classic curriculum learning consolidates knowledge from observed
tasks and delivers a more consistent performance across all the tasks as
compared to the baseline. However, the method still results in catastrophic forgetting
(Fig.~\ref{fig:training_curves}), as the cues learned from solving smaller problem sizes
are not helpful enough to the model to retain that knowledge. One way to
alleviate forgetting is to combine rehearsing with incremental learning.

We propose to use adaptive staircase, a method previously applied
to train a reinforcement learning agent in a psychophysics environment \citep{leibo2018psychlab}.
The experiment was run by scheduling tasks according to the probe case with $\alpha=0.05$
threshold on the optimality gap to advance the task difficulty. If performance $\alpha$
was not attained after a full epoch we decremented the task. We set the number of base trials
(batches) per epoch to
\textit{(10+current\_graph\_size - min\_graph\_size) / (max\_graph\_size - min\_graph\_size)}.
To make sure that the number of probe trials did not exceed the number of batches available to
the other sampling strategies, we stopped training one epoch earlier than the other strategies. This
resulted in $\approx 1.6\%$ reduction in the amount of data exposed to this strategy.
Fig.~\ref{fig:training_curves} and Fig.~\ref{fig:all_methods_opt_gap} both show
that adaptive staircase results in the best trade-off between forgetting and
average performance across problem sizes.

\section{Conclusion}

In this work we have assessed the ability of an attention-based NCO model to perform well
on a large range of problem sizes. We
found that knowledge transfer between tasks is measurable and changes
smoothly, but is otherwise limited and cannot be ignored
when designing a sampling strategy. Curriculum learning provides a convenient
framework that takes into account complexity and knowledge transfer between
tasks. The classic approach of training on tasks monotonically increasing in
complexity suffers from catastrophic forgetting. Our analysis demonstrates the
need for rehearsal of less complex samples during training. By choosing an efficient
rehearsal-based sampling strategy, adaptive staircase, we were able to show both the
usefulness of curriculum learning in neural combinatorial optimization and provide
important evidence supporting the applicability of a biologically-inspired method to an emerging area of deep learning.

\begin{ack}
The authors would like to thank Vithursan Thangarasa for preliminary discussions. We also thank Irwan Bello from Google Brain as well as our lab colleagues, Magdalena Sobol and Angus Galloway, for their support in reviewing this manuscript.
\end{ack}

\bibliographystyle{plainnat}
\bibliography{references}

\end{document}